\RequirePackage{fix-cm}

\documentclass[twocolumn]{svjour3}          % twocolumn
\smartqed  % flush right qed marks, e.g. at end of proof
\usepackage{graphicx}
\usepackage{amsmath}
\usepackage{algpseudocode}
\usepackage{algorithm}
\usepackage{slashbox}
\usepackage{multirow}

\usepackage{mathtools}

\journalname{Technical Report}

\begin{document}

\title{%Sign Language and Action Classification \\based on Body Parts Relations
Reasoning about Body-Parts Relations \\for Sign Language Recognition.
% % % \thanks{This work was supported by the FP7 ERC grant 240530 COGNIMUND, the KU
% % % Leuven OT Project VASI and the DBOF Research Fund KUL 3E100864.}
% 
\thanks{This work is partially supported by FWO project G.0.398.11.N.10 ``Multi-camera human behavior monitoring and unusual event detection'',  KU Leuven GOA project CAMETRON, the ``Fondo Europeo de Desarrollo Regional'' (FEDER) and the Spanish
Ministry of Economy and Competitiveness, under its R\&D\&i Support Program in project with ref TEC2013-45492-R.}
% 
% \thanks{Grants or other notes
}


\author{
	Marc Mart{\'i}nez Camarena\textbf{$^*$} \and
	Jos{\'e} Oramas M.\textbf{$^*$}         \and
	\\Mario Montagud Climent \and
        Tinne Tuytelaars       %etc.
}


\institute{Marc Mart{\'i}nez Camarena, Mario Montagud Climent, \at 
	   IMM, Universidad Polit{\'e}cnica de Valencia (UPV), Spain.
	   \and
           Jos{\'e} Oramas M., Tinne Tuytelaars \at
           KU Leuven, ESAT-PSI, iMinds Belgium.
           \and
           \textbf{*} First two authors had an equal contribution to this work.
% % %               Tel.: +123-45-678910\\
% % %               Fax: +123-45-678910\\
% % %               \email{fauthor@example.com}           %  \\
}

\date{Received: date / Accepted: date}

\maketitle

\begin{abstract}
Over the years, hand gesture recognition has been mostly addressed considering 
hand trajectories in isolation. However, in most sign languages, hand gestures
are defined on a particular context (body region). 
We propose a pipeline to perform sign language recognition which models hand 
movements in the
context of other parts of the body captured in the 3D space using the MS Kinect sensor. 
In addition, we perform sign recognition based on the different 
hand postures that occur during a sign. 
Our experiments show that considering different body parts brings improved performance 
when compared to other methods which only consider global hand trajectories. 
Finally, we demonstrate that the combination of hand postures features with 
hand gestures features helps to improve the prediction of a given sign.

\keywords{Hand gesture recognition \and sign language recognition \and 
relational learning \and classification.}
\end{abstract}

\begin{figure}[t]
	\centering
		\includegraphics[width=0.44\textwidth]{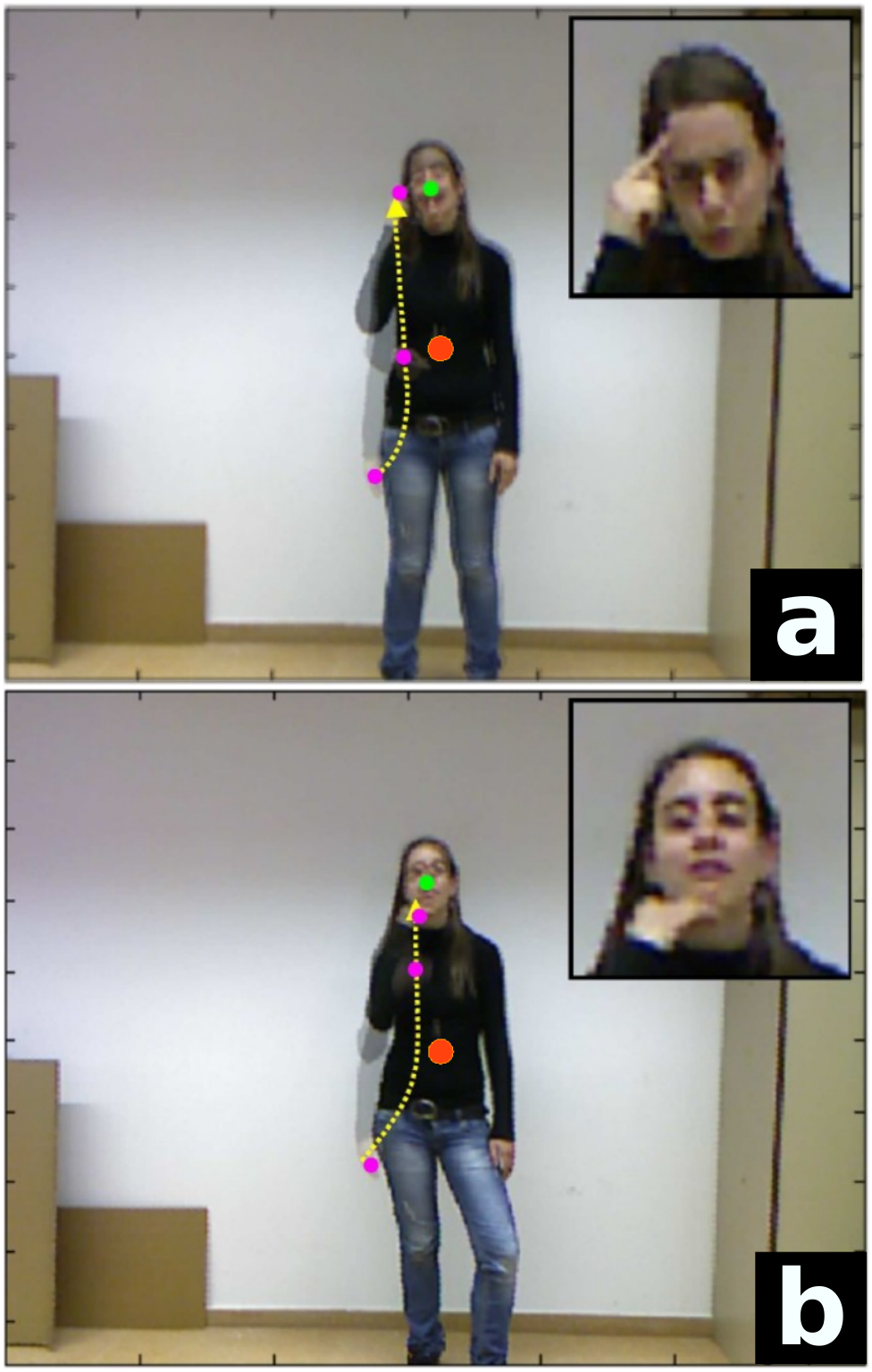}
		\caption{Note how signs with similar global trajectories (in yellow) can be distinguished based
		    on the relative locations of the hand (in magenta) w.r.t. the head (in green). 
		    In addition, see how the posture of the hands can help to distinguish between 
		    similar signs (see insets). Selected body part locations in color. Green: head location, 
		    magenta: right hand, orange: torso. 
		    Images taken from the ChaLearn Gestures dataset \cite{EscaleraICMI13}. 
		    (Best viewed in color).}
		
\vspace*{-0.3cm}
\label{fig:problemStatement}
\end{figure}

\section{Introduction}
Hearing-impaired people as a community consider themselves 
a minority who communicates differently 
rather than a group of disabled people. Unfortunately, in some 
countries, this minority faces difficulties during their 
teaching/learning process. One of the most critical factors is 
the low teacher-student ratio, which directly affects the learning 
of communication skills by young students. As a consequence 
this hampers the possibility of the student for self-learning. 
Hence there is a clear need for a system to learn/practice sign language.

There is a wide variety of sign languages that are used by 
hearing-impaired individuals around the world. Each language is formed by 
grammar rules and a vocabulary of signs. Something that most of 
these languages have in common is that signs are composed 
by two elements: \textit{hand postures}, i.e. the position or 
configuration of the fingers; and \textit{hand gestures}, 
i.e. the movement of the hand as a whole. 
In this paper we focus on the problem of sign classification 
based on hand postures and hand gestures, leaving 
elements such as facial gestures or grammar rules for future work.

Initial work on sign language recognition has been based 
on sensor gloves \cite{DipietroGloveSystems} or shape descriptors 
for the recognition of postures. For the recognition of gestures 
accelerometers or colored gloves have been used to assist tracking 
the hand. 
In recent years the release of the MS Kinect device, a low-cost 
depth camera, has provided means to acquire relatively accurate 
3D data about objects. This has been followed by a variety of 
prototype gestural interfaces and definitively gives an 
opportunity to provide an automatic solution that can alleviate the problem
related to low teacher-student ratio, previously introduced.
However most of these prototype gesture interfaces consider 
very simple gestures, mostly targeted to interaction with consumer products, 
or employ weak gesture description methods that are not 
suited to accurately recognize sign language.

In this work, we consider relations between 
different parts of the body for the task 
of sign language recognition. For example, see how  
the global motion of the sign in 
Figure~\ref{fig:problemStatement}(a) is very similar to the
motion of the sign in Figure~\ref{fig:problemStatement}(b). 
However, the relative motion of the hand~(magenta) w.r.t. 
the head~(green) is different for both signs, especially 
at the very end.
Our method uses a first generation MS Kinect device to capture 
the data, in particular RGBD images to localize the 
different body parts. 
Then, each sign is represented by a 
combination of responses obtained from cues extracted 
from hand postures and hand gestures, respectively.
For the problem of sign language recognition 
based on hand posture cues, we use  shape context descriptors 
in combination with a multiclass Support Vector Machine (SVM) classifier to recognize 
the different signs. Regarding sign recognition based on cues 
derived from hand gestures, we use Hidden Markov Models (HMMs) 
to model the dynamics of each gesture.  
Finally, sign prediction is achieved by the late fusion of the responses 
of the processes for sign recognition based on hand postures and 
gestures, respectively.
This paper extends our previous work \cite{MartinezICIP15} 
in four directions. First, we provide an extended discussion of 
related work, taking into account the recent literature. Second, 
we provide a more detailed presentation of the internals of our 
method, complemented by related experiments. Third, we propose 
and evaluate an alternative method for fusing the responses based 
on hand postures and gestures features, respectively. Finally, 
we extend our evaluation to two additional datasets.

The main contribution of this work is to show that reasoning 
about relations between parts of the body for the recognition
of hand gestures brings improvements for hand gesture recognition 
and has potential for sign language recognition. 
This paper is organized as follows: 
Section~\ref{sec:relatedWork} positions our work with 
respect to similar work. In Sections~\ref{sec:proposedMethod} and 
\ref{sec:implementationDetails} we present the details of our 
method and its implementation, respectively. Section~\ref{sec:evaluation} 
presents the evaluation protocol and experimental results.
In Section~\ref{sec:conclusion} we conclude this paper.

%%%%%%%%%%%%%%%%%%%%%%%%%%%%%%%%%%%%%%%%%%%%%%%%%%%%%%%%%%%%%%%%%%%%%%%%%%%%%%%%
\section{Related Work}
\label{sec:relatedWork}

For many years, the art of gesture recognition has been 
mostly focused on 2D information \cite{ArgyrosECCV04,joramasEMinds11,YuWCSE09}. 
However, using this approach, there are still several 
challenges to be addressed, for instance, illumination 
change, background clutter, etc. Recently, with the advent 
of low-cost depth-cameras, reasoning can be focused 
in 3D space (e.g. \cite{BillietICPRAM13,KuzLea2013}), 
using jointly depth and color images. 
Working in the 3D space the problems of illumination 
change and background clutter can be reduced. In 
addition, the objects of interest can be isolated or 
segmented more accurately. Thanks to the recent 
development of inexpensive depth cameras, we will 
adopt a low-cost vision-based approach in which we 
use the consumer camera Kinect. Starting from this 
point, existing work can be divided into the four 
following groups:

\subsection{Hands-focused Methods}
Previous works in vision-based sign language recognition 
have mostly focused on isolating the hands and then reasoning 
about features extracted exclusively from them.
These works formulate the sign language recognition 
problem either as a hand posture recognition problem or as 
a trajectory matching problem.
For instance, for the case of hand posture-based methods, 
In \cite{ThangaliCVPR11}, a non-rigid image alignment algorithm 
is proposed to enforce robustness towards hand shape variations. 
Furthermore, a Bayesian network formulation is used to enforce 
linguistic constrains between the hand shapes at start and end 
of a sign. 
In \cite{RenTMM13}, Ren et al. propose a novel distance metric 
using RGBD images to measure the dissimilarity between hand 
shapes. Their method is able to distinguish slightly-different 
hand postures since they match the finger parts rather than 
the whole hand. Similarly, in \cite{KuzLea2013}, depth images 
are used to extract rotation, translation and scale invariant 
features which are used to train a multi-layered random forest 
model. This model is later used to classify a newly observed 
hand posture. Billiet et al.~\cite{BillietICPRAM13} 
present a model-based approach in which they represent 
the hand using pre-defined rules. Their hand model is 
based on a fixed number of hand components. Each component 
is a finger group with its associated finger pose. 
The hand is segmented based on depth information. Then, 
the RGB image is used to recognize the different hand 
postures.
For the case of trajectory-based methods, a common practice is
to track and describe the global motion of the hand, either in 
the 2D image space or the 3D scene space. In \cite{joramasEMinds11}
color filtering in the HSV space is used to segment the hands in 
the image space. Then, the global 2D trajectories of the hands 
are represented as regular expressions and matched against a 
set of pre-defined rules representing hand gestures of interest. 
In \cite{ChaiFG13}, RGBD images collected with 
a first-generation MS Kinect are used to estimate the 3D location 
of the hands. During testing, recognition is achieved by aligning 
the global 3D motion trajectory of a given sign w.r.t. each sign 
from a pre-defined vocabulary of signs.
More recently, Wang et al. \cite{Wang3DTrajectory} proposed a method 
where signs are described by
typical posture fragments, where hand motions are relatively slow and 
hand shapes are stable. In addition, the 3D motion trajectory of each 
hand is integrated taking into account the position and size of the 
signer. During testing, the sequence of hand postures and the 3D 
trajectories are matched against a gallery of sign templates.
These methods achieve good results, however they focus their 
reasoning on features derived from the hands in isolation. 
Compared to these works, our method takes into account the 
context (parts of the body) in which the hand trajectories occur. 
In addition, for the works that rely on modeling hand postures, 
their methods rely on an accurate construction of a hand model. 
However, such accurate construction may not be possible for the 
case of low-resolution images as is typically the case when one 
wants to extract gesture-based information at the same time. 
On the contrary, we propose a method based on lower-level 
features which relaxes the requirement of high-resolution images.

% \begin{figure}[htbp]
\begin{figure*}[ht!]
	\centering
		\includegraphics[width=0.7\textwidth]{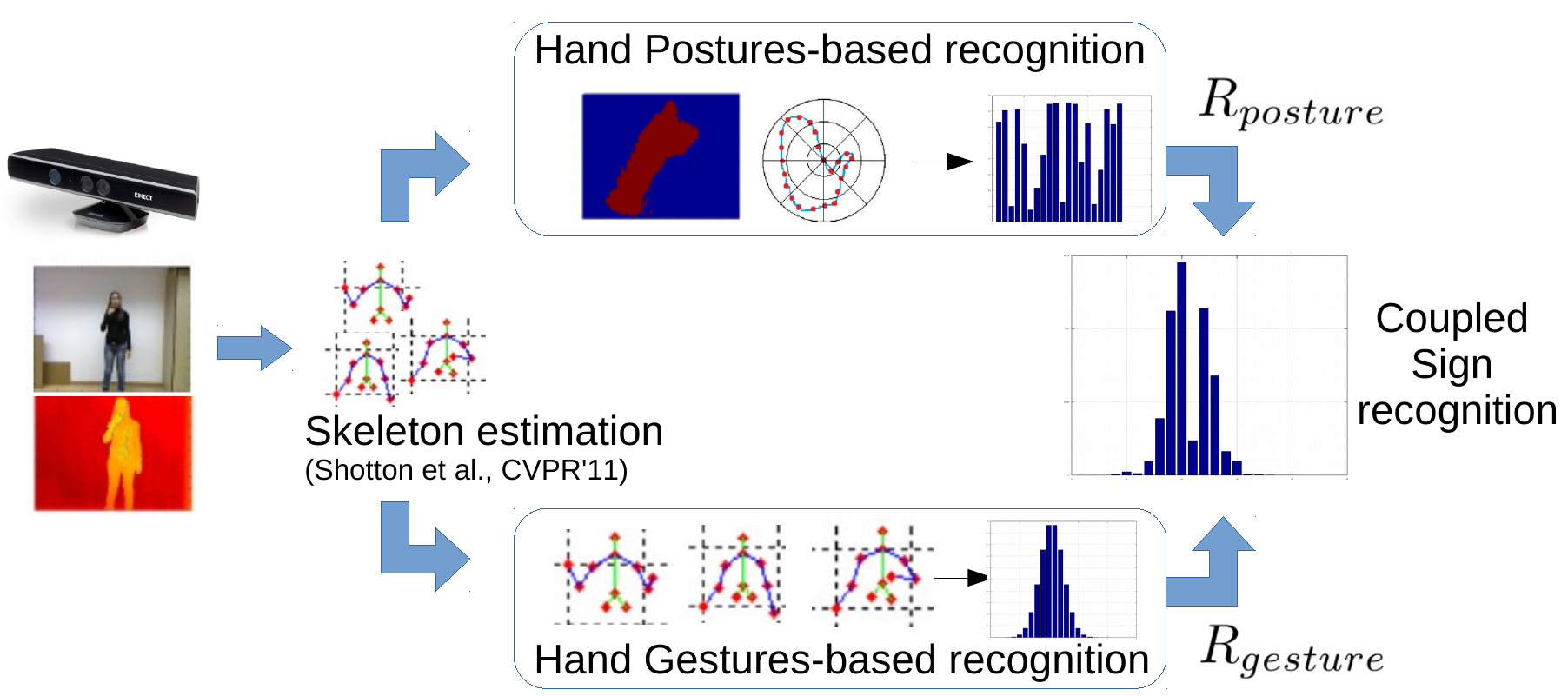}
		\caption{Algorithm pipeline. Skeleton joints are estimated using a MS 
		         Kinect and the method from \cite{ShottonCVPR11}. Then, the 
		         distribution over all possible sign classes is computed based 
		         on posture and gesture features, respectively. Finally, these 
		         two responses are combined and the final sign label is predicted.}
		         
\vspace*{-0.3cm}		
\label{fig:overviewsystem}
\end{figure*}

\subsection{Exploiting skeleton representations}
Regarding hand gesture recognition, skeleton-based algorithms 
make use of 3D information to identify key elements, in particular 
the human body parts. 
A milestone method for the extraction of the human body 
skeleton is presented by Shotton et al. \cite{ShottonCVPR11}. Ever since, 
the human skeleton model has been widely used for gesture recognition 
since this approach allows relatively accurate tracking of the joints
of the body in real-time.  
Papadopoulo et al. \cite{Papadopoulos13} use the skeleton representation 
to compute the joint angles and angular velocities between each pair 
of connected parts. Then, these descriptors are used to identify action poses, 
such as: clapping, throwing, punching, etc. In \cite{WuICMI2013}, 
a 12-dimensional skeleton-based feature vector is defined by considering 
global 3D location of four joints of the skeleton (left/right elbow and 
left/right wrists). During testing, the label of an unknown sequence is 
estimated by measuring its similarity w.r.t. training sequences via Dynamic 
Time Warping (DTW) \cite{Muller07}. 
In \cite{EllisIJCV2013}, body pose information is encoded by 
computing the pairwise distances between 15 joints. In parallel, 
motion information is encoded by computing the Euclidean 
distance between pairs of joints detected at the current frame 
and joints detected 10 frames earlier. In addition, to encode 
overall dynamics of body movement,  similar pairwise distances 
are computed between the current frames and a frame where the 
person is in a resting position. Finally, by using GentleBoost, 
the most discriminative features are identified and used for 
testing.
Similar to these works, we also use an implementation of the algorithm 
from \cite{ShottonCVPR11} to acquire  the set of points of the 
skeleton in each frame and build a descriptor modeling the joints 
of the hands with respect to the joints of the other parts of the body. 

\subsection{Mid-level Representations for Gesture Recognition}
Sign/Gesture recognition can be approached by performing 
classification directly from features computed on shapes (postures) 
or trajectories (gestures) done with the hands. However, there is 
a more recent trend in which these initial features are used to 
define mid-level representations. These representations are 
general enough to be used as a common vocabulary along different 
action/gesture classes. Futhermore, they can cope with small 
intra-class variations that can be introduced by different individuals 
performing the actions/gestures.
Following this trend, Ellis et al. \cite{EllisIJCV2013} propose a 
Logistic Regression learning framework that automatically finds the 
most discriminative canonical body pose representation of each action 
and then performs classification using these extracted poses. 
In \cite{HusseinIJCAI2013}, the covariance matrix between skeleton 
joint locations over time is used as a descriptor (Cov3DJ) for a sequence.
The relationship between joint movement and time is encoded by 
taking into account multiple covariance matrices over sub-sequences 
in a hierarchical fashion. Labeled training data is encoded with 
these descriptors and a linear SVM classifier is trained which is 
later used during testing. 
More recently, in \cite{PonceBMVC2015}, a set of shared 
spatio-temporal primitives, subgestures, are detected using 
genetic algorithms. Then, the dynamics of the actions of 
interest are modeled using the detected primitives and either 
HMMs or DTW. Similar to the previous works 
we use mid-level representations to perform hand gesture 
classification. In this work, we first compute pairwise relations 
between skeleton joints for each frame in our training sequences.
Then, we re-encode each of the sets of pairwise relations via 
K-Means, where each cluster center is a representative pose that 
the body can take when performing one of the gestures/actions. 

% \begin{figure}[htbp]
\begin{figure*}
% \begin{figure}[t]
	\centering
		\includegraphics[width=0.98\textwidth]{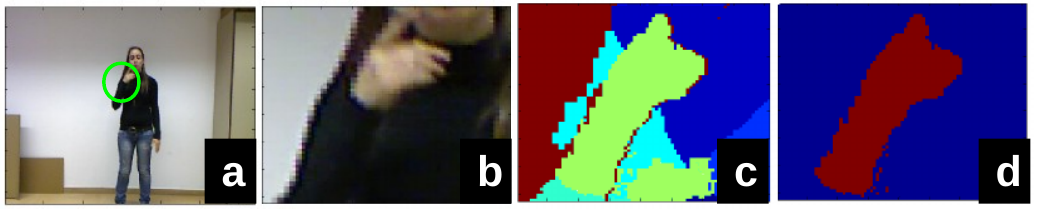}
		\caption[Steps followed to isolate the hand]{ Hand segmentation algorithm: 
		(a) Original RGB image collected with kinect,
		(b) Cropped RGB image after spatial thresholding,
		(c) Projected 3D points assigned to the different parts of the body 
		    (light green:hand, cyan:arm, blue:shoulder, red:background), and
	        (d) hand region $H$ after applying binarization to the 2D points derived from 
	            the 3D points assigned to the parts of the hands. The largest region
	            assigned the hand label (in red) is selected.}
\label{fig:stepsextracthandSVMs}
\end{figure*}

\subsection{Modeling Hand Gestures Dynamics}
Apart from the spatial representation of hand gestures, another main problem
to be solved is the temporal alignment among different sequences. Hand 
gestures may be understood as continuous sequences of data points or temporal
series. The most modern approaches include Dynamic Time Warping (DTW),
Conditional Random Fields (CRF), Hidden Markov Models (HMMs) and Rank Pooling. 
Despite several extensions of DTW, the disadvantage of using DTW 
is the heavy computational cost involved to find the optimal time 
alignment path, which makes DTW practical only for small data sets. 
CRF is based on discriminative learning.
In \cite{ChungOE13} Chung and Yang use a  CRF with a threshold model to 
recognize the different feature vectors which are described by the angular 
relationship between body components in 3D space. Different from CRF, HMM 
is a generative method which learns how to model each class independently 
of the rest.  In \cite{Elmezain09}, Elmezain et al. propose a system to 
recognize the alphabet (American Sign Language) and numbers in real time 
by tracking the hand trajectory using HMMs. Gu et al. \cite{Gu13} 
implement a gesture recognition system using the 3D skeleton provided by 
the MS Kinect device. They use HMMs to model the dynamics of the training 
gestures, one HMM per gesture class.
Very recently, Fernando et al.~\cite{FernandoAlCVPR15} proposed Rank Pooling,
a method to perform action recognition by modeling the evolution of 
frame-level features over time by using ranking machines. In the context 
of gesture recognition, in \cite{FernandoAlCVPR15} the skeleton joints 
are computed using the method from \cite{ShottonCVPR11}. Then, for each 
frame, the relative location of each body joint w.r.t. the torso joint 
is computed (similar to our HD baseline, see Section~\ref{sec:exp:gestureRecog}). 
Each frame is re-encoded using the learned parameters of a ranking machine 
trained to order these skeleton quantized features chronologically. Finally, 
a SVM classifier is trained and used later during test time.
In this work, we have chosen HMMs due to their remarkable performance 
on gesture recognition, being used in the top performing methods in 
previous editions of the ChaLearn Gesture Challenge~\cite{EscaleraICMI13,GuyonICPR13}. 
In the proposed method, HMMs are used to model the sequential transition 
between body poses acquired with Kinect during each of the signs of interest. 
This allows us to perform sign recognition based on cues derived 
from relative gestures of the hands.

%%%%%%%%%%%%%%%%%%%%%%%%%%%%%%%%%%%%%%%%%%%%%%%%%%%%%%%%%%%%%%%%%%%%%%%%%%%%%%%%
\section{Proposed Method}
\label{sec:proposedMethod}

The proposed method can be summarized in the following steps 
(see Figure~\ref{fig:overviewsystem}): First, a MS Kinect device 
is used to capture the RGB and depth images. Based on these images 
we estimate the skeleton body representation using the algorithm from 
Shotton et al.~\cite{ShottonCVPR11}. Then, our method consists of two parallel 
stages: the recognition of signs based on hand posture features and the 
recognition of signs based on hand gesture features. 
Finally, the response of the recognition of signs based on hand 
posture features is combined with the response based on gesture features 
to estimate the likelihood of a given sign.

\subsection{Sign Recognition based on Hand Postures}
\label{sec:postures}

\subsubsection{Hand Region Segmentation}
\label{sec:postureSegmentation}
The component based on hand posture features takes as input RGBD images
and the skeleton body representation estimated using a MS Kinect device 
in combination with the algorithm from Shotton et al. \cite{ShottonCVPR11}. 
In order to segment the hand region, the 3D world coordinate space is calculated 
from the depth images obtaining the $(X,Y,Z)$ coordinates of all the points 
of the scene. To reduce the number of points to be processed, we perform 
an early spatial threshold to filter points far from the expected hand 
regions. To this end, all the points outside the sphere centered on the 
hand joint whose radius is half the distance between the joints of the hand 
and elbow are removed (Figure~\ref{fig:stepsextracthandSVMs}(a,b)).
Once the amount of points has been reduced, we assign the remaining 3D
points to the closest body joint, estimated via \cite{ShottonCVPR11}, using
Nearest Neighbors~(NN) classification (Figure~\ref{fig:stepsextracthandSVMs}(c)).
This cluster assignment is computed in the 3D space, keeping correspondences 
with the pixels in the image space. Following the cluster assignment, we
only keep the points that were assigned to the joints of the hands.
For the case of multiple regions assigned to the hand joint, we keep
the largest region. This allows our method to overcome noise introduced
by low resolution images and scenarios in which the hand comes in contact
with other parts of the body.
In addition, we re-scale the depth images to a common 65x65 pixels patch.
Finally, as Figure~\ref{fig:stepsextracthandSVMs}(d) shows, we binarize 
the re-scaled patch producing the hand regions $H$ . 
Figure~\ref{fig:stepsextracthandSVMs} shows the different steps starting 
from the input RGBD image until obtaining the hand region $H$.

% \begin{figure}[htbp]
\begin{figure}
	\centering
		\includegraphics[width=0.5\textwidth]{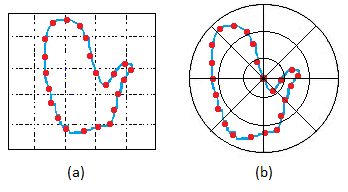}
		\caption{Computation of Shape Context descriptors: (a) Selection of equally-spaced points on the hand region $H$ contour,
		         and (b) Log polar sampling (8 angular and 3 distance bins).}
	\label{fig:shapecontext_des}
% % % 	\vspace{-3mm}
\end{figure}

\subsubsection{Hand Posture Description}
\label{sec:postureDescription}

A side effect of capturing full-body images is 
that they result in hand regions which lack details 
(Figure~\ref{fig:stepsextracthandSVMs}(b)). Hence, 
a method to robustly encode information from 
low-resolution hand regions from images is desirable.
For this reason, 
once we have obtained the candidate 2D regions $H$ 
containing the hands, we describe the different hand 
postures by a Bag-of-Words representation constructed 
from shape context descriptors~\cite{BelongieSC}.

In order to compute the shape context descriptor $s$, we extract a number of
$m$ equally-spaced points from the contour of each binary hand 
region $H$ (Figure~\ref{fig:shapecontext_des}(a)) obtained from the
hand segmentation step. Then, using this set of points, a log-polar binning 
coordinate system is centered at each of the points and a 
histogram accumulates the amount of contour points that fall within 
each bin (Figure~\ref{fig:shapecontext_des}(b)).
This histogram is the shape context descriptor.
This procedure is performed on each frame of the video sequence. Then,
we define a Bag-of-Words representation $p$ where each video is a bag containing a set 
of words from a dictionary obtained by vector-quantizing the shape context descriptors 
$s$ via K-means. 
This procedure is applied for both hands of the user producing two descriptors, 
$(p_{right}$, $p_{left})$, one for each hand, which are concatenated into one
posture-based descriptor $p=[p_{right},~p_{left}]$.

\subsubsection{Recognizing signs based on hand postures features}
\label{sec:postureClassification}
Once the posture descriptors $p_i$ for all the MS Kinect video 
sequences have been computed, we train a multiclass SVM classifier using 
the pairs $(p_i,c_i)$ composed by the concatenated 
posture-based descriptor $p_i$ with its corresponding sign class $c_i$. 
We follow a one-vs-all strategy and the method from Crammer and Singer~\cite{crammerJMLR2001} 
to train the classifier and learn the model $W$. 

During testing, given a video sequence captured with MS Kinect, a similar 
approach is followed to obtain the representation $p_i$ based on posture features.
Then, as Eq.~\ref{eq:SVMposture} shows, the learned model $W$ is used to 
compute the response $R_{posture}$ of the input video sequence over the 
difference sign classes, based purely on hand postures.

\begin{equation}
%     \vspace*{-0.05cm}
    \label{eq:SVMposture}
%     \hat{c_i} = \arg_{c_k} \max ( W_k \cdot R_i).
     R_{posture} =  W * p_i.
\end{equation}

where $p_i$ is the posture-based descriptor computed from the testing 
example and $W = [W_1,W_2,...,W_k]^T$ is the matrix of weights from the
SVM models (one for each of the sign classes), purely based on hand posture 
features.

\subsection{Sign Recognition based on Hand Gestures}
\label{sec:gestures}

\begin{figure}
% % \begin{figure}[htbp]
	\centering
		\includegraphics[width=0.25\textwidth]{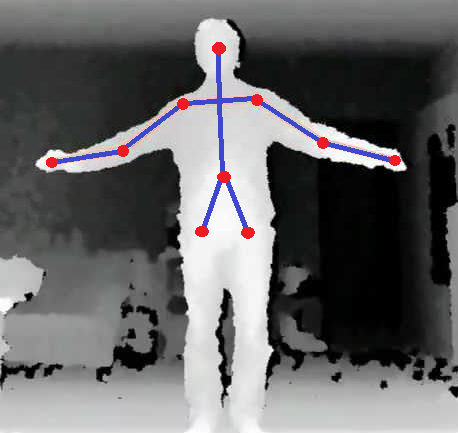}
		
% % % % 		\caption{The new coordinate systems defined by each hand for the computation of  \textit{RBPD}.}
		\caption{Skeleton joints of the upper part of the body considered for describing signs.}
	\label{fig:skeleton11Joints}
\end{figure}

Similarly to the hand posture component (Section~\ref{sec:postures}), 
we take as input for the hand gestures component RGBD images collected with kinect and
the human skeleton joints estimated using the method from \cite{ShottonCVPR11}.
The goal of this component is to infer from this skeleton a set of features that
enable effective recognition of signs based on hand gestures. Towards this goal,
from the initial set of 15 3D joints,  we only 
consider a set $J=\{j_1,j_2,...,j_{11}\}$ of 11 3D joints covering the upper body 
(see Figure~\ref{fig:skeleton11Joints}). This is due to the fact that most of the 
sign languages only use the upper part of the body to define their signs. 

\begin{figure}
% % \begin{figure}[htbp]
	\centering
		\includegraphics[width=0.46\textwidth]{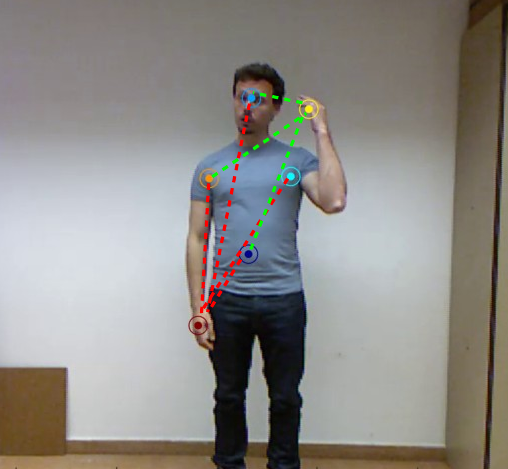}

		\caption{Relative Body Parts Descriptor (\textit{RBPD}) computation. For clarity, we only show
		         the \textit{RBPD} computed for head, torso and shoulder joints for the left (green) 
		         and right (red) hands, respectively. In practice, these descriptors are computed between 
		         the 3D locations of the hands wrt. all the joints of the upper part of the body.}
	\label{fig:skeleton_2_CS}
\end{figure}

\subsubsection{Hand Gesture Representation}
 \label{'RBPDdescriptor'}

 \begin{figure*}[t]
	\centering
% % % 		\includegraphics[width=1\textwidth]{pictures/skeletonVisualWords.png}
		\includegraphics[width=0.96\textwidth]{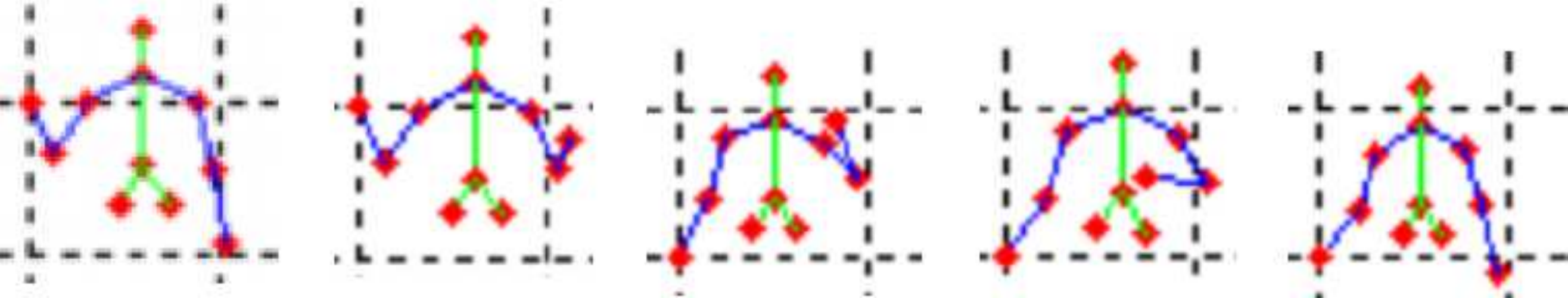}
		\caption{Examples of cluster centers in the set of relative body poses from the whole 
		         training data of the ChaLearn dataset~\cite{EscaleraICMI13}.}
% % % \vspace*{-0.3cm}
\label{fig:skeletonWords}
\end{figure*}
 
% % % \vspace{5mm}
% % % \textbf{Relative Body Part Descriptor:} 
Once the set of joints $J$ have been selected, we define a descriptor to
represent hand gestures based on relations between the hands and the
rest of joints, or parts, of the body.
This is motivated by two observations: first, because most sign languages 
use hands as the main, or most active, element of the signer. Second, because
during different hand gestures the hands may follow similar trajectories, 
however these trajectories can be defined in the context of different body areas.
For example, in Figure~\ref{fig:problemStatement}, even when the signs in 
row 1 and row 2 have a similar global trajectory (Figure~\ref{fig:problemStatement}.(a)), 
in yellow, the sign in row 1 involves hand contact on top of the head, 
while the sign in row 2 involves contact with the lower part of the 
head (Figure~\ref{fig:problemStatement}.(b)).

Given the set $J$ of selected joints where each joint $j=(X,Y,Z)$ is 
defined by its 3D location. We define the Relative Body Part 
Descriptor~(RBPD) as $RBPD=[\delta_1,\delta_2,...,\delta_m]$ where 
$\delta_i = (j_i-j_h)$ is the relative location of each non-hand joint 
$j_i$ w.r.t. one of the hand joints $j_h$ (Figure~\ref{fig:skeleton_2_CS}).
We perform this operation for each of the two hands. The final
descriptor is defined by the concatenation of the descriptors computed
from each hand $RBPD=[RBPD_{right},RBPD_{left}]$. Notice that the length 
of this descriptor is 66 since we are considering 11 parts of the body 
including the hands. 
Furthermore, note that, the user can be at different locations 
with respect to the visual field of the camera and consequently there 
might be considerable variation in  $X ,Y$  and $Z$ coordinates. 
However, by building the proposed descriptor, considering relative 
locations between the hands w.r.t. body, we achieve some level 
of invariance towards translation in the location of the user. 
Finally, until now, the estimated input 
descriptor \textit{RBPD} constitutes the observation at a specific frame. 
In order to extend this frame-level representation to the full 
gesture sequence we compute this descriptor for each of the $n$ 
frames of the video $g=[RBPD_{1}, RBPD_{2}, ..., RBPD_{n}]$.

% % % \vspace{5mm}
% \textbf{Visual Dictionary:}
% % % \textit{Visual Dictionary:}
\subsubsection{Mid-level Feature Encoding}
\label{sec:bodyposedictionary}

Up to this point, every video sequence is represented as a sequence of 
displacement vectors between body parts ($RBPD$). Each vector 
being computed independently of the user and the sign class. However, even 
when focusing on single sign class, different users may introduce small 
variations to the sign they perform. Likewise, some sign classes may share 
some characteristics at the gesture level. To address these issues, we 
re-encode the $RBPD$s by using a mid-level representation that 
can be shared between both users and sign classes. To this end, we compute 
the $RPBD$s from all the frames of the training sequences, z-normalize them, 
and cluster them using K-means with. This $K$ value was obtained from 
running the pipeline in the validation set. Then, each video is re-encoded 
by the sequence of cluster centers $w_i$ that its corresponding $RPBD$s are 
assigned to.
As a result, each gesture will now be represented by a sequence of centers $w$.
These cluster centers $w$ are stored for later use during the testing stage.
See Figure~\ref{fig:skeletonWords} for some examples of the cluster centers 
$w$.

% % % \subsubsection{ Recognizing signs based on hand gesture features}
% % \textbf{ Recognizing signs based on hand gesture features:}
% % % \vspace{5mm}
% % % \noindent\textit{ Recognizing signs based on hand gesture features:}
\subsubsection{Recognizing signs based on hand gesture features}
\label{sec:gestureClassification}
In this paper, we model the dynamics of the hand gestures using left-right 
Hidden Markov models (HMMs). Specifically, we train one HMM per sign class.
HMMs are a type of statistical model which are characterized by the number 
of states in the model, the number of distinct observation symbols per state,
the state transition probability distribution, the observation symbol 
probability distribution and the initial state distribution. 
In our system, the training observations 
$(o_{1}, o_{2}, ..., o_{n})$ are the  hand gestures represented as 
a sequence of centers estimated from the encoding step. 
These observations $o_{i}$ are collected per sign class $c_i$ and used
to train each HMM.  
The state transition probability of each model
is initialized with the value 0.5 to allow each state to begin 
or stay on itself with the same probability.
The number of states is different for each model and was determined using 
validation data. In addition, the number of distinct observation symbols of
the models is equal to the number of centers $K$.
Furthermore, in order to ensure that the models begin from their 
respective first state, the initial state distribution gives all 
the weight to the first state.
Finally, the observation symbol probability distribution matrix
of each model is uniformly initialized with the value $1/K$, where
$K$ is the number of distinct observation symbols.
During training, for each model, the state transition probability
distribution, the observation symbol probability distribution and 
the initial state distribution are re-adjusted by using the Baum-Welch 
algorithm~\cite{JelinekBaumWelchAlgorithm}.
Once the different HMMs have been trained for each sign 
class $c_i$, the system is then ready for sign classification.
During testing, given a gesture observation $g$, sequence of encoded
centers, and a set of pre-trained HMMs
$\Omega$, our method selects the class of the model $\Omega_k$ that 
maximizes the likelihood $p(c|g)$ of class $c$ based on gestures features
~(Eq.\ref{eq:posetureSignClassification}).
In this paper we refer to such likelihood $p(c|g)$ as the sign response
$R_{gesture}$ based, purely, on hand gesture features.

\begin{equation}
  c = \arg_k \max p(k|g) = \arg_k \max (\Omega_k(g))
  \label{eq:posetureSignClassification} 
\end{equation}

\subsection{Coupled Sign Language Recognition}
\label{sec:coupleSignRecognition}
For each RGBD sequence captured with MS kinect, the ealier
components of the system compute the $R_{posture}$
and $R_{gesture}$ responses over the sign classes based on posture
and gesture features, respectively. 
In order to obtain a final prediction, we define the coupled
response $R$ by late fusion of the responses $R_{posture}$
and $R_{gesture}$. To this end, given a set of validation sequences,
for each example sequence we compute the responses based 
on the postures $R_{posture}$ and gestures $R_{gesture}$. 
In addition, for each example, we define the coupled 
descriptor $R=[R_{posture} , R_{gesture}]$ as the 
concatenation of the two responses. 
Then, using the coupled descriptors - class 
label pairs~$(R_i,c_i)$ from each validation example
we train a multiclass SVM classifier using linear kernels.
This effectively learns the optimal linear combination of
$R_{posture}$ and $R_{gesture}$.
During testing, the sign class $\hat{c_i}$ is obtained as: 

\begin{equation} 
%     \vspace*{-0.05cm}
    \label{eq:SVMcomb}
    \hat{c_i} = \arg_{c_k} \max ( \omega_k \cdot R_i).
\end{equation}

where $R_i$ is the coupled response computed from the testing 
data and $\omega = [\omega_1,\omega_2,...,\omega_k]^T$ are the weight 
vectors from the SVM models.

\paragraph{}
In addition to the previous SVM-based method to perform a linear combination
of the responses, we explore the performance of an alternative probabilistic 
method \cite{PerkoCVIU10} to combine the responses. 
Given the coupled descriptors - class label pairs~$(R_i,c_i)$ 
from each validation example, the sign class $\hat{c_i}$ or $R_i$ is the 
MAP estimate by applying the Bayes rule:

\begin{equation} 
    \label{eq:ODKDEcomb}
    \hat{c_i} = \arg_{c_k} \max p(R_i|c_k)p(c_k),
\end{equation}

where the class likelihoods $p(R_i|c_k )$ are computed using Kernel Density
Estimation (KDE) and the priors $p(c_k)$ are obtained from the occurrence of  
sign class $c_k$ on the validation data.

%%%%%%%%%%%%%%%%%%%%%%%%%%%%%%%%%%%%%%%%%%%%%%%%%%%%%%%%%%%%%%%%%%%%%%%%%%%%%%%%
\section{Implementation Details}
\label{sec:implementationDetails}

In this section we provide some implementation details in order to ease the 
reproducibility of the method proposed in this paper.

%%% POSTURES 
As mentioned in Section~\ref{sec:postureDescription} we define our posture-based representation
from shape context descriptors \cite{BelongieSC}. In our experiments, we use a pseudo 
log-polar sampling mask with 12 angular and 5 distance bins (with an inner radius 6 
pixels and an outer radius 32 pixels) delivering a 60 dimensional histogram for each of 
the sampled points. We combine the inner part of the log polar mask used to build 
the shape context descriptor into one bin since there is some evidence \cite{Mikolajczyk2005}
that combining this inner part produces improved results. 
For this reason, the length of our shape context descriptor is reduced to 49 dimensions. 
Then, each shape context descriptor is normalized dividing each element of the 
descriptor by the sum all the elements of the descriptor. 
Once the hand region $H$ has been segmented, the shape descriptor is computed on a 
total of 20 equally spaced points.
When performing K-means a value of K=100 was used since that value gave the best 
performance in the validation set.
In addition, during SVM training (Section~\ref{sec:postureClassification}), at the 
posture stage, we use 3-fold cross validation and a cost value $C=0.8352$.
%%% GESTURES
% For the construction of the visual dictionary based on hand gestures 
% (Section~\ref{sec:bodyposedictionary}), we use a value of $K=95$ for K-means.

%%% COUPLED
During the coupled sign recognition stage, Section~\ref{sec:coupleSignRecognition}, 
we train the SVM models via 3-fold cross validation with a cost value $C=0.7641$.
%%% SVM
In our implementation, we use the Liblinear~\cite{FanLiblinear} for SVM training 
and classification. We perform multiclass classification following a one-vs-all 
strategy and the method from Crammer and Singer~\cite{crammerJMLR2001} to train 
the models. 
%%% ODKDE
For the case of the alternative probabilistic response-fusion method based on KDE, 
we use the Online Kernel Density Estimation (oKDE) variant proposed in 
\cite{KristanLODKDE14,KristanOKDE11}. However, since no online learning/estimation 
is required, we apply low compression and construct the initial estimator from 
the whole set of training examples. In consequence, we only keep its variable 
multivariate properties for kernel density estimation.

%%%%%%%%%%%%%%%%%%%%%%%%%%%%%%%%%%%%%%%%%%%%%%%%%%%%%%%%%%%%%%%%%%%%%%%%%%%%%%%%
\section{Evaluation}
\label{sec:evaluation}
In this section, we present the experimental protocol followed to 
evaluate the performance of the proposed method. We divide our 
evaluation into five subsections aimed at analyzing different 
aspects of the proposed method. To this end, we evaluate its performance when 
only considering posture-based features (Section~\ref{sec:exp:postureRecog}), its performance when only considering gestures-based features (Section~\ref{sec:exp:gestureRecog}), the combination of both 
posture and gesture features (Section~\ref{sec:exp:coupledRecog}), a comparison of the proposed method w.r.t. state-of-the-art methods (Section~\ref{sec:exp:comparison}), and its computation time (Section~\ref{sec:exp:compTime}).

%%% CHALEARN-2013
We evaluate our approach on the ChaLearn Multi-modal Gesture
Recognition Challenge 2013. This dataset was introduced in 
\cite{EscaleraICMI13}. It contains 20 Italian cultural / anthropological 
signs produced by a total of 27 subjects. It provides RGBD images 
captured with a MS Kinect device plus the skeleton joints estimated using 
the method from \cite{ShottonCVPR11}.
All the examples in the training and validation sets are annotated at frame level indicating the beginning and ending frame of each sign. Since the focus of our work is towards sign classification rather than sign detection, we organize our data in isolated sign sequences.
For the sake of comparison with recent work 
\cite{PfisterECCV14,WuICMI2013,Yao_2014_CVPR}, we use the original training set of the dataset for training and the original validation set of the dataset for testing. This is split in such a way that ensures that a subject whose data occurs in the training set, does not occur in the testing set. In addition, we split the training set into two subsets, one subset for training and one subset for validation purposes. 
Moreover, different from the original, Levenshtein distance, performance metric used in the challenge~\cite{EscaleraICMI13}, we report results using as performance metric mean precision, recall and F-Score.
In addition, for reference, we present results on the original testing set of the ChaLearn Gesture dataset which annotations were kindly provided by its organizers.

%%% MSR Action3D dataset
Additionally, we also conduct experiments on the MSR Action3D dataset~\cite{Li_actionrecognition}. 
It includes 20 classes of actions. Each action was performed by 10 subjects 
for three times. This dataset was captured at 15 fps with a resolution of 320x240. 
It is composed by 23797 frames of RGBD images for 402 action sequences. 
For the sake of comparison we follow a similar evaluation protocol as proposed in 
\cite{Li_actionrecognition,WangCVPR12} to split the data into training and testing 
sets. We report performance in terms of mean accuracy, precision, recall and F-Score.
Different from the ChaLearn Gestures dataset, this dataset is more oriented towards 
general actions, e.g. ``pick up and throw'', ``golf swing'', ``hand clap'', 
``hammer'', etc. However, we will only focus only on the joints of the upper part 
of the body for the description of hand gestures.

%%% MSRC-12 dataset
Finally, we perform experiments on the MSRC-12 dataset \cite{msrc12}.
This dataset is captured at 30 fps and composed of 594 sequences (719359 frames) 
from 30 subjects performing 12 gestures.
We conduct experiments on the MSRC-12 dataset, following the protocol 
from \cite{EllisIJCV2013,HusseinIJCAI2013}. Different from the ChaLearn dataset, 
the MSRC-12 dataset does not include RGBD images for the sake of anonymity. 
For this reason, we cannot report results for the combined method on MSRC-12, since 
RGBD images are required for the processing of hand postures.

\begin{figure*}
	\centering
		\includegraphics[width=0.72\textwidth]{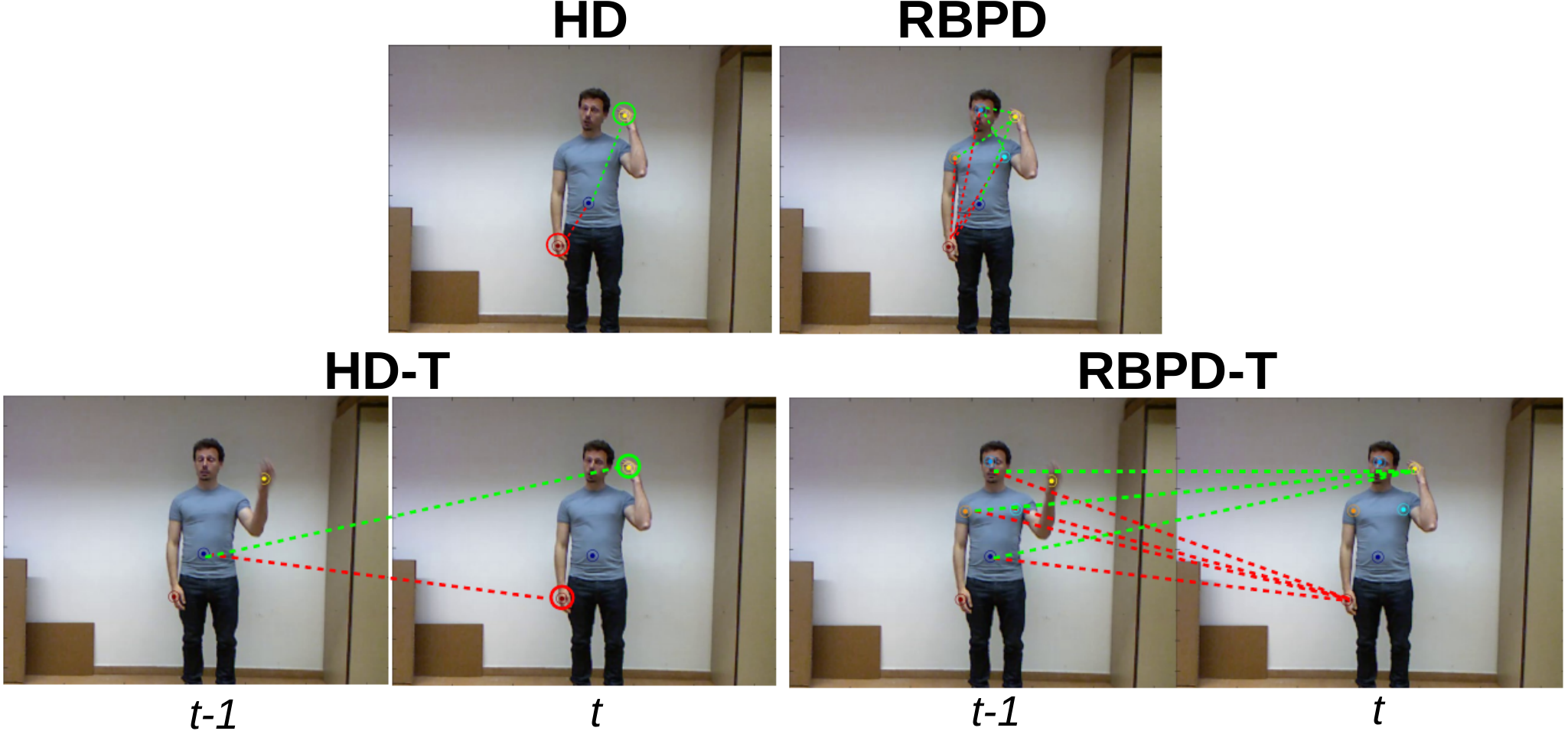}
		\caption{Evaluated methods to model hand gestures. The purely spatial descriptors 
		which operate at the frame level: hand descriptor (\textit{HD}) and Relative 
		Body Parts Descriptors (\textit{RBPD}), and their time-extesion counterparts, 
		\textit{HD-T} and \textit{RBPD-T}, which operate between frames at different 
		time stamps.}
% % % \vspace*{-0.3cm}
\label{fig:gestureDescriptors}
\end{figure*}

\subsection{Sign Recognition based on Hand Postures}
\label{sec:exp:postureRecog}

In this experiment we evaluate the performance of the method at
recognizing signs based purely on features derived from hand postures.
Table~\ref{table:postureBasedPerfomanceTable} presents the mean performance 
of our method when only considering postures computed from hand postures 
(Section~\ref{sec:postures}). Figure~\ref{fig:confusionMatrices}(first column) 
shows the confusion matrix of this experiment in the test set. 
% % % Tables~\ref{tabletodoPrec}, \ref{tabletodo} and \ref{tabletodoFScore} 
% % % show further results at the class level.

\begin{table}[ht]
\caption{Hand Posture-based recognition mean performance.}
\centering

\begin{tabular}{|c|c|c|c|}
\hline
\multicolumn{3}{|c|}{\textbf{ChaLearn (val.) dataset \cite{EscaleraICMI13}}} \\ \hline
Precision & Recall & F-Score \\ \hline
0.42 & 0.35 & 0.38    \\ \hline

\multicolumn{3}{c}{} \\ \hline

\multicolumn{3}{|c|}{\textbf{ChaLearn (test.) dataset \cite{EscaleraICMI13}}} \\ \hline
Precision & Recall & F-Score \\ \hline
0.34 & 0.33 & 0.34    \\ \hline

\multicolumn{3}{c}{} \\ \hline

\multicolumn{3}{|c|}{\textbf{MSR Action3D dataset \cite{Li_actionrecognition}}} \\ \hline
Precision & Recall & F-Score \\ \hline
0.40 & 0.40 & 0.40    \\ \hline

\end{tabular}
\label{table:postureBasedPerfomanceTable}
\end{table}

\textit{Discussion:}
recognition based on hand posture features has an average F-Score of $0.38$ , 
$0.34$ and $0.40$, on the ChaLearn (validation and testing sets) 
and MSR Action3D datasets, respectively.
This low average is due to the fact that: (1) the signs were captured at a 
distance around two to three meters from the camera obtaining images with poor 
resolution, specially for the regions that cover the hands. In addition, it 
should be noted that the hand, compared with the complete human body, is a 
smaller deformable object and more easily affected by segmentation error. 
(2) On many of the signs, the hands come into contact or get very 
close to the body (see Figure~\ref{fig:problemStatement}(b)) making it 
difficult to obtain a good segmentation and introducing error in the 
features computed from the hand region. 
(3) Some of the signs are defined with very similar sequences of hand 
postures, being only different in one or two hand postures along the 
sequence (in this case, the most significant hand posture(s) that define
the sign) easily leading to sign miss classification. 
If we compare our shape context-based method with other methods for hand 
posture recognition \cite{BillietICPRAM13,KeskinICCVWS2011,RenTMM13}, 
we notice that our method is better suited for these low-resolution images. 
This is due to the fact that our method does not rely on the 
construction of a more detailed hand model which is a difficult task 
on low-resolution images like the ones of the ChaLearn gestures dataset. 
On the contrary, our method is able to leverage posture features from 
low-resolution images removing the requirement of a detailed hand model.

\begin{figure*}[t]
	\centering
		\includegraphics[width=1\textwidth]{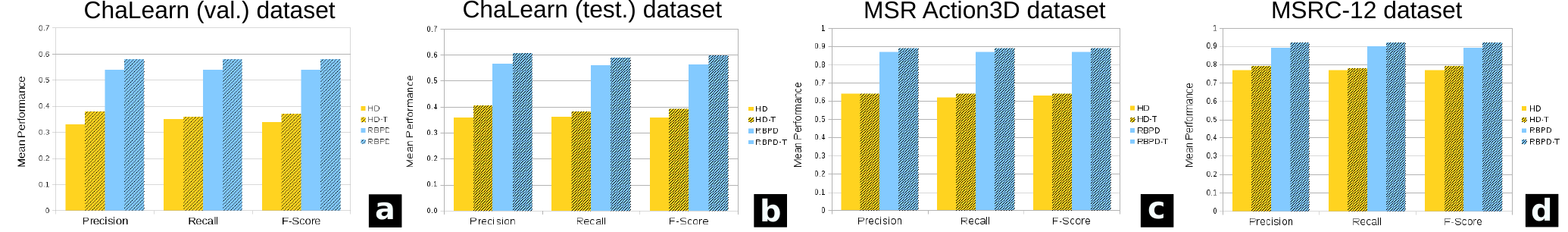}
		\caption{Gesture-based recognition mean performance on the validation (a) and test (b) set of the ChaLearn gestures dataset~\cite{EscaleraICMI13}, the MSR Action3D dataset (c) \cite{Li_actionrecognition}, and the MSRC-12 dataset (d) \cite{msrc12}. Note how the performance of purely focusing on global hand trajectories (\textit{HD, HD-T}), presented in yellow color, 
		is much lower than the performance of our method considering body part relations (\textit{RBPD, RBPD-T}), presented in light blue color. Furthermore, note how considering the time extension of the descriptors (\textit{HD-T,RBPD-T}) brings a small improvement over their purely spatial counterparts (\textit{HD,RBPD}).
		         }
\vspace*{-0.4cm}
\label{fig:gestureBasedPerfomancePlot}
\end{figure*}

\subsection{Sign Recognition based on Hand Gestures}
\label{sec:exp:gestureRecog}
\begin{figure*}
	\centering
		\includegraphics[width=1\textwidth]{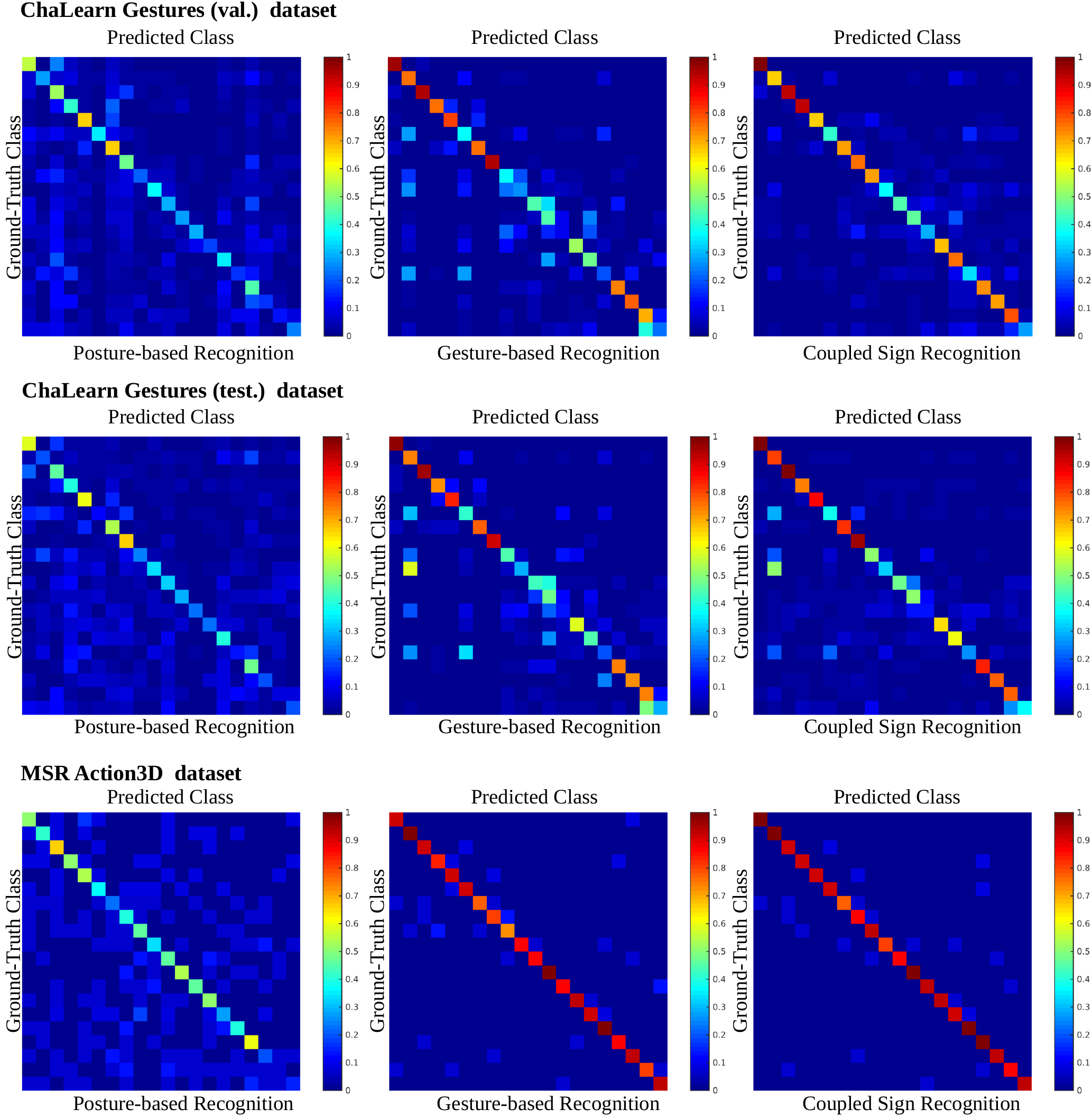}
		\caption{Confusion matrices for sign recognition based on responses computed from Hand Postures (first column), Hand Gestures (second column) (\textit{RBPD-T}), and late fusion (probabilistic) of hand postures and gestures responses (third column).}
% % % \vspace*{-0.3cm}
\label{fig:confusionMatrices}
\end{figure*}

\vspace*{-0.2cm}
In this experiment we focus on the recognition of signs 
based on hand gestures. We evaluate
four methods (see Figure~\ref{fig:gestureDescriptors}) 
to model the gestures: a) the \textit{RBPD} 
descriptor proposed in Section~\ref{sec:gestures}; 
b) the \textit{RBPD-T} descriptor which is similar to 
\textit{RBPD}, however, in this descriptor the relations 
between the hands and the other parts of the body are 
estimated taking into account the hand locations in the 
current frame and the location of the other parts in the 
next frame. As a result, this descriptor not only takes 
into account spatial relations but implicitly adds temporal 
features; c) the \textit{HD} descriptor which only 
considers the location of the hands w.r.t. the torso 
location; and d) \textit{HD-T}, a time extension of \textit{HD}.
The last two methods, \textit{HD} and \textit{HD-T}, 
are based on hand trajectories since we only follow 
the location of the hands over time. 
Similar to \textit{RBPD}, we train HMMs (Section ~\ref{sec:gestureClassification}) 
using these methods, \textit{RBPD-T}, \textit{HD}, 
and \textit{HD-T}, for gesture representation.
% % % In the validation set, these three methods achieved mean F-Score
% % % performance of $0.56, 0.62$ and  $0.36$, respectively.
From these methods, we take the top performing \textit{RBPD-T} 
for further experiments. 
Figure~\ref{fig:gestureBasedPerfomancePlot} and 
Table~\ref{table:gestureBasedPerfomanceTable} show the mean 
performance of each of these methods to model gestures in the 
evaluated datasets.
Figure~\ref{fig:confusionMatrices}(second column) shows the 
confusion matrix of recognizing signs based on hand gesture 
features.

% % % % % % % % % % % % % % % % % % % % % % % % % % % % % % % % % % % % % % % % % % % 

\begin{table}
\caption{Gesture-based recognition mean performance.}
\centering

\begin{tabular}{|l |c|c|c|}
\hline
& \multicolumn{3}{|c|}{\textbf{ChaLearn (val.) dataset \cite{EscaleraICMI13}}} \\ \cline{2-4}
& Precision & Recall & F-Score \\ \hline
HD     & 0.33 & 0.35 & 0.34    \\ \hline
HD-T   & 0.38 & 0.36 & 0.37    \\ \hline
RBPD   & 0.54 & 0.54 & 0.54    \\ \hline
RBPD-T & \textbf{0.58} & \textbf{0.58} & \textbf{0.58}  \\ \hline
% % % \multicolumn{4}{l}{20 classes}\\

\multicolumn{4}{c}{} \\ \hline

& \multicolumn{3}{|c|}{\textbf{ChaLearn (test) dataset \cite{EscaleraICMI13}}} \\ \cline{2-4}
& Precision & Recall & F-Score \\ \hline
HD     & 0.36 & 0.36 & 0.36    \\ \hline
HD-T   & 0.40 & 0.38 & 0.39    \\ \hline
RBPD   & 0.57 & 0.56 & 0.56    \\ \hline
RBPD-T & \textbf{0.61} & \textbf{0.59} & \textbf{0.60}  \\ \hline
% % % \multicolumn{4}{l}{20 classes}\\

\multicolumn{4}{c}{} \\ \hline

& \multicolumn{3}{|c|}{\textbf{MSR Action3D dataset \cite{Li_actionrecognition}}} \\ \cline{2-4}
& Precision & Recall & F-Score  \\ \hline

HD     & 0.64 & 0.62 & 0.63 \\ \hline
HD-T   & 0.64 & 0.64 & 0.64 \\ \hline
RBPD   & 0.87 & 0.86 & 0.87 \\ \hline
RBPD-T & \textbf{0.89} & \textbf{0.89} & \textbf{0.89}  \\ \hline

\multicolumn{4}{c}{} \\ \hline

& \multicolumn{3}{|c|}{\textbf{MSRC-12 dataset \cite{msrc12}}} \\ \cline{2-4}
& Precision & Recall & F-Score  \\ \hline

HD     & 0.77 & 0.77 & 0.77 \\ \hline
HD-T   & 0.79 & 0.78 & 0.79 \\ \hline
RBPD   & 0.89 & 0.90 & 0.89 \\ \hline
RBPD-T & \textbf{0.92} & \textbf{0.92} & \textbf{0.92}  \\ \hline

\end{tabular}
\label{table:gestureBasedPerfomanceTable}
\end{table}

% % % % % % % % % % % % % % % % % % % % % % % % % % % % % % % % % % % % % % % % % % % % % 

% % % % % % % % % % % % % % % % % % % % % % % % % % % % % % % % % % % % % % % % % % % 

% % % % % % % % % % % % % % % % % % % % % % % % % % % % % % % % % % % % % % % % % % % % % d

\begin{figure*}[t]
	\centering
		\includegraphics[width=1\textwidth]{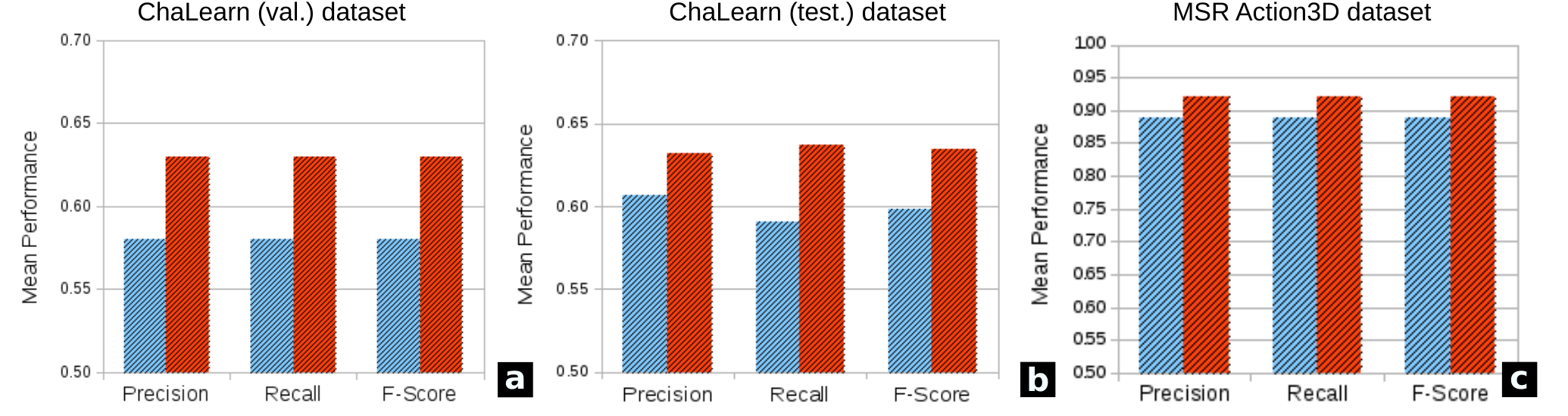}
		\caption{Mean F-Score when performing sign recognition when considering only gesture-based 
		features (light blue) and when considering both hand postures and hand gestures features 
		(orange) on the ChaLearn~\cite{EscaleraICMI13} (a,b) and MSR Action3D~\cite{Li_actionrecognition} (c)
		datasets.}
% % % \vspace*{-0.3cm}
\label{fig:coupledRecognitionPerfomancePlot}
\end{figure*}

% % % % % % % % % % % % % % % % % % % % % % % % % % % % % % % % % % % % % % % % % % % 

\begin{figure}
	\centering
		\includegraphics[width=0.49\textwidth]{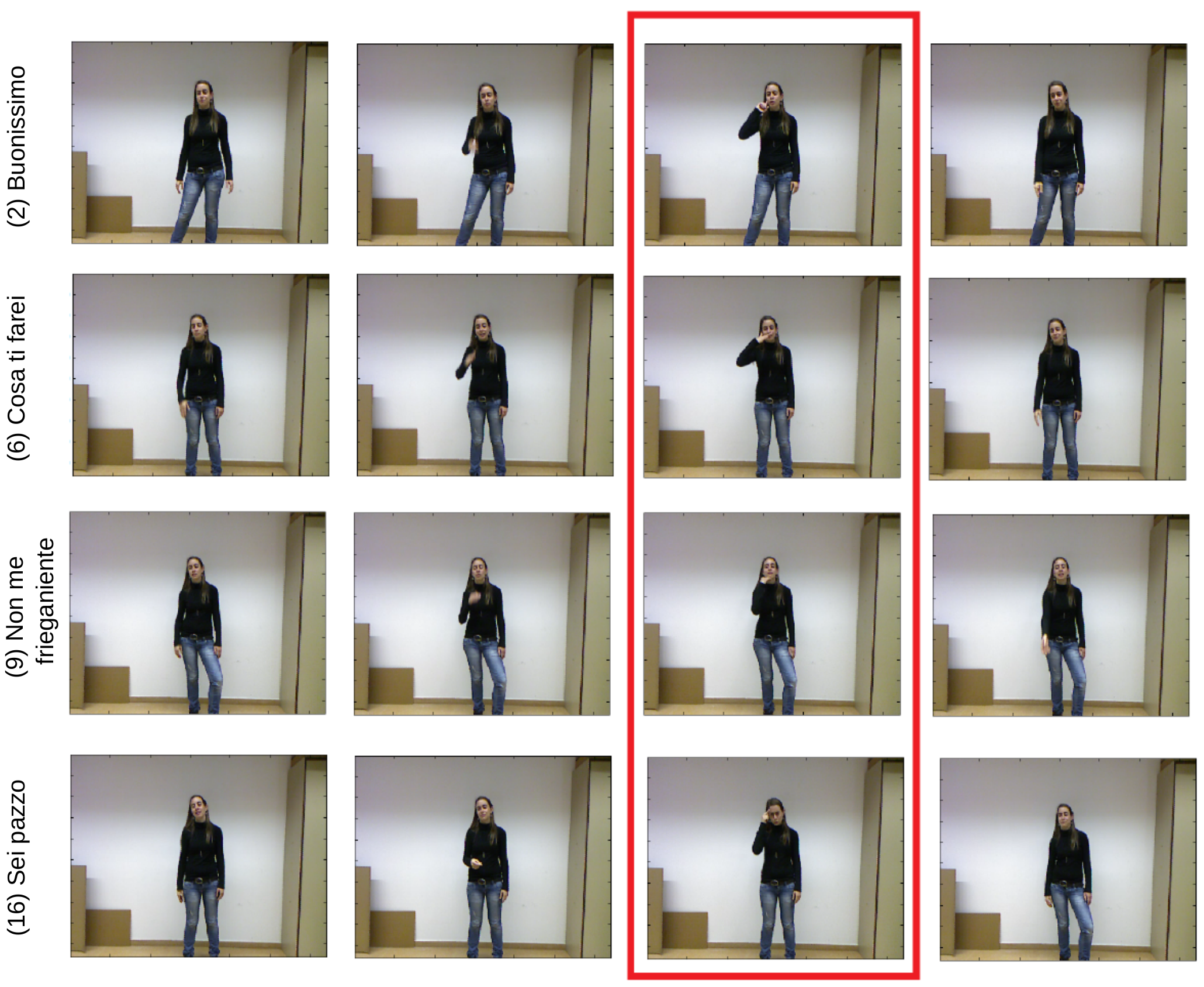}
		\caption{Some of the confusing signs from the ChaLearn gestures dataset \cite{EscaleraICMI13} 
		when only considering gesture-based information. Notice how the motion of the hand is very 
		similar along the different signs. However, they can still be differentiated by the posture 
		of the hand (marked by the red box).}
% % % \vspace*{-0.3cm}
\label{fig:confusingGestureClassesChalearn}
\end{figure}

% % % \subparagraph{\textbf{Discussion:}}
% \textbf{Discussion:}
\textit{Discussion:}
A quick inspection to Figure~\ref{fig:gestureBasedPerfomancePlot} reveals that 
taking into account relations between different body parts when modeling hand 
gestures brings improvements over methods that only consider global hand 
trajectories for sign recognition.
This is supported by an improvement of 24 percentage points (pp) higher mean 
F-Score of \textit{RBPD} over \textit{HD} on the ChaLearn Gestures and MSRC-12 
datasets, Note that these datasets are more sign language oriented. For the case 
of the MSR Action3D dataset, this improvement of performance is around 24
pp, still confirming that this collective reasoning about parts of the body 
brings improvement to action recognition.
Compared to this, the differences between the performance of the \textit{HD,RBPD} 
and \textit{HD-T,RBPD-T}, resepctively, seem to be minimal. 
The time extensions \textit{HD-T} and \textit{RBPD-T} seem to bring higher 
improvement the closer the problem is to a sign language recognition setting.
For example, in the ChaLearn dataset, it brings an improvement of $\sim$4 pp
while in the MSRC-12 dataset this improvements drops to $\sim$2 pp. This further 
drops to $\sim$1 for the more general action classes of the the MSR Action3D dataset 
(see Table~\ref{table:gestureBasedPerfomanceTable}).
As Figure~\ref{fig:confusionMatrices}(second column) shows the \textit{RBPD-T} is able to 
recognize some signs with very high accuracy, e.g. signs 1, 3 and 8 of the ChaLearn dataset. This is due to the fact that these signs are more different from 
the other signs such that their respective hand gesture representations,
sequence of cluster centers, are more unique. As Figure~\ref{fig:confusionMatrices}(second column) also shows the main problems are the confusions of the signs 6, 9, 10 and 16 with 2 and signs 11, 13 and 15 with sign 12 due to the similarities between 
the gestures of those signs being differentiated mostly by particular 
hand postures (see Figure~\ref{fig:confusingGestureClassesChalearn} for a qualitative example). For the case of the MSR Action3D dataset Figure~\ref{fig:confusionMatrices}(second column, last row) , where hand postures take a secondary role, we can notice that hand gestures-based features alone can produce very good performance.

% % % % % % % % % % % % % % % % % % % % % % % % % % % % % % % % % % 

% % % % % % % % % % % % % % % % % % % % % % % % % % % % % % % % % % % % % % % % % % % 

\subsection{Coupled Sign Recognition}
\label{sec:exp:coupledRecog}

In this experiment we evaluate the performance of the coupled
response $R=[R_{posture} , R_{gesture}]$ based on hand posture 
and hand gesture features as described in Section~\ref{sec:coupleSignRecognition}.
We compare the performance provided by the two methods presented 
in Section~\ref{sec:coupleSignRecognition} to perform the combination 
of the responses based on hand postures and hand gestures features, 
respectively.
Table~\ref{table:coupledCombinationMethods} presents the performance 
of different response combination methods on the ChaLearn gestures dataset 
and the MSR Action3D dataset. 
% % % % In addition, for reference, we present results on the testing set of the ChaLearn dataset which annotations were kindly provided by its organizers. 
As mentioned earlier, no performance on 
Coupled Sign Recognition is presented for the MSRC-12 dataset since no 
hand posture information can be extracted from it.
Figure~\ref{fig:confusionMatrices}(third column) shows the confusion matrix for the combination of the responses in the evaluated datasets.

% % % % % % % % % % % % % % % % % % % % % % % % % % % % % % % % % % % % % % % % % % % 

\begin{table}
\caption{Coupled recognition mean performance. Gestures features are based on \textit{RBPD-T}.}
\centering

\begin{tabular}{|l |c|c|c|}
\hline
& \multicolumn{3}{|c|}{\textbf{ChaLearn (val.) dataset \cite{EscaleraICMI13}}} \\ \cline{2-4}
Fusion Method & Precision & Recall & F-Score \\ \hline
Linear Combination & 0.61 & 0.62 & 0.62    \\ \hline
Probabilistic & \textbf{0.63} & \textbf{0.63} & \textbf{0.63}  \\ \hline

\multicolumn{4}{c}{} \\ \hline

& \multicolumn{3}{|c|}{\textbf{ChaLearn (test.) dataset \cite{EscaleraICMI13}}} \\ \cline{2-4}
Fusion Method & Precision & Recall & F-Score \\ \hline
Linear Combination   & \textbf{0.63} & 0.62 & 0.62    \\ \hline
Probabilistic 	     & \textbf{0.63} & \textbf{0.64} & \textbf{0.63}  \\ \hline

\multicolumn{4}{c}{} \\ \hline

& \multicolumn{3}{|c|}{\textbf{MSR Action3D dataset \cite{Li_actionrecognition}}} \\ \cline{2-4}
Fusion Method & Precision & Recall & F-Score \\ \hline
Linear Combination     & 0.91 & 0.91 & 0.91    \\ \hline
Probabilistic & \textbf{0.92} & \textbf{0.92} & \textbf{0.92}  \\ \hline

\end{tabular}
\label{table:coupledCombinationMethods}
\end{table}

% % % % % % % % % % % % % % % % % % % % % % % % % % % % % % % % % % % % % % % % % % % % % % % % % % % % % % % % % % % % % % % 

% % % \subparagraph{ \textbf{Discussion:}}
% \textbf{Discussion:}
\textit{Discussion:} 
At first sight, as Figure~\ref{fig:coupledRecognitionPerfomancePlot} 
shows, the combination of responses, based on hand postures and 
gestures features, outperforms the overall performance of the method 
when considering only hand gestures.
In addition, it can be noted from the confusion matrices (Figure~\ref{fig:confusionMatrices}) 
of both methods that confusion between sign classes is reduced 
showing the complementarity of both responses, based on postures 
and gestures, respectively. This is to be expected since some 
ambiguous cases can be clarified by looking at the relations
between parts of the body (see row 1 vs row 2 of 
Figure~\ref{fig:problemStatement}(a)). Likewise, other ambiguous 
cases can be clarified by giving more attention to 
the hand postures (Figure~\ref{fig:confusingGestureClassesChalearn} third column).
In addition, we can notice in Table~\ref{table:coupledCombinationMethods}
that the proposed methods to combine the responses based on hand postures 
and hand gestures have a similar performance. Nevertheless, the probabilistic 
method based on KDE provides an improvement $\sim$1 pp over the method 
based on linear combination of the responses.

\subsection{Comparison w.r.t the state-of-the-art}
\label{sec:exp:comparison}

Given the observations made in the previous experiments, 
in this experiment we select the top-performing method, i.e. 
\textit{RBPD-T} for gesture modeling and probabilistic 
combination of responses, and used it for comparison w.r.t.  
the state-of-the-art.
For the case of the ChaLearn gestures (val.) dataset we compare with 
recent work \cite{FernandoAlCVPR15,PfisterECCV14,MartinezICIP15,WuICMI2013,Yao_2014_CVPR}. 
We report results in Table~\ref{table:comparisonLiterature_chalearn} 
using as performance metric mean precision, recall and F-Score.
Similarly, for the MSRC-12 dataset, we compare with 
\cite{EllisIJCV2013,HusseinIJCAI2013}.
For the case of the MSR Action3D dataset, we follow the 
evaluation protocol from \cite{Li_actionrecognition,JetleyC14,WangLCCW12,WangCVPR12} 
and compare the performance of our method with the one reported 
by those methods, respectively. Table~\ref{table:comparisonLiterature_MSRAction3D} 
reports the results in terms of Mean Accuracy.

% % % % % % % % % % % % % % % % % % % % % % % % % % % % % % % % % % % % % % % % % % % 

\begin{table}
\caption{Comparison with the State of the Art in chronological order.
Mean performance over all the 20 sign classes of the ChaLearn 2013 dataset~\cite{EscaleraICMI13}.}
\centering
\begin{tabular}{|l |c|c|c|}
\hline
& Precision & Recall & F-Score  \\ \hline

Wu et al.,~\cite{WuICMI2013}        & 0.60 & 0.59 & 0.60  \\ \hline
Yao et al.,~\cite{Yao_2014_CVPR}     &  -   &   -  & 0.56  \\ \hline
Pfister et al.,~\cite{PfisterECCV14} & 0.61 & 0.62 & 0.62  \\ \hline
Fernando et al.,~\cite{FernandoAlCVPR15} & \textbf{0.75} & \textbf{0.75} &
\textbf{0.75}  \\ \hline
Ours (linear comb.) &  0.61 & 0.62 & 0.62  \\ \hline
Ours (probabilistic comb.) &  0.63 & 0.63 & 0.63  \\ \hline
\end{tabular}
\vspace{-2mm}
\label{table:comparisonLiterature_chalearn}
\end{table}

% % % % % % % % % % % % % % % % % % % % % % % % % % % % % % % % % % % % % % % % % % % % % % 

\begin{table}
\caption{Comparison with the State of the Art in chronological order.
Mean Accuracy over all the 20 classes of the MSR Action3D 2013 dataset~\cite{Li_actionrecognition}.}
\centering
\begin{tabular}{|l|c|}
\hline
& Accuracy \\ \hline

Li et al.,~\cite{Li_actionrecognition}          & 0.747 \\ \hline 
Wang et al.,~\cite{WangCVPR12}      		& 0.882 \\ \hline
Wang et al.,~\cite{WangLCCW12}			& 0.862 \\ \hline	
Ellis et al.,~\cite{EllisIJCV2013}         	& 0.657  \\ \hline
Hussein et al.,~\cite{HusseinIJCAI2013}   	& 0.905  \\ \hline
Jetley et al.,~\cite{JetleyC14}   		& 0.838 \\ \hline
Ponce-L{\'o}pez et al.,~\cite{PonceBMVC2015}   	& \textbf{0.950} \\ \hline
Ours (linear comb.)				& 0.908 \\ \hline
Ours (probabilistic comb.) 			& 0.919 \\ \hline
\end{tabular}
\label{table:comparisonLiterature_MSRAction3D}
\end{table}

% % % % % % % % % % % % % % % % % % % % % % % % % % % % % % % % % % % % % % % % % % % % 

\begin{table}
\caption{Comparison with the State of the Art in chronological order.
Mean Accuracy over all the 12 gesture classes of the MSRC-12 dataset~\cite{msrc12}.}
\centering
\begin{tabular}{|l |c|}
\hline
& Accuracy  \\ \hline

Ellis et al.,~\cite{EllisIJCV2013}       & 0.887  \\ \hline
Hussein et al.,~\cite{HusseinIJCAI2013}  & 0.903  \\ \hline
Ours (RBPD-T)  				 &  \textbf{0.919}   \\\hline

\end{tabular}
\vspace{-2mm}
\label{table:comparisonLiterature_msrc12}
\end{table}

% % % % % % % % % % % % % % % % % % % % % % % % % % % % % % % % % % % % % % % % % % % % 

\textit{Discussion:}
% ChaLearn dataset
Compared to \cite{WuICMI2013}, the method that was ranked 1st in the 
Multi-modal Gesture Recognition Challenge in 2013 \cite{EscaleraICMI13}
(when only using image/video data), our combined method achieves an improvement 
of $\sim$4 F-Score pp over their method (Table~\ref{table:comparisonLiterature_chalearn}).
Furthermore our method is still superior by $\sim$7 pp over the F-Score 
performance reported by the recent method from \cite{Yao_2014_CVPR}.
This is to be expected since our method explicitly exploits information
about hand postures, which \cite{Yao_2014_CVPR} ignores. This last
feature makes the proposed method more suitable to address sign language
recognition where hand posture information is of interest. Even more, our
method has a comparable performance (1 pp improvement on performance) to 
the method from \cite{PfisterECCV14}, which also considers hand posture 
information. However, different from \cite{PfisterECCV14}, our method does 
not rely on face detection and skin segmentation in order to localize the 
hand regions. Compared to the just-published method from \cite{FernandoAlCVPR15}, 
our method achieves inferior performance ($\sim$13 pp lower F-Score). 
The method from \cite{PfisterECCV14} uses hand trajectories and the 
method from \cite{BuehlerIJCV11,BuehlerCVPR09} for hand posture modeling. 
This is closer to the \textit{HD} method that we evaluated which, in our 
experiments, produced suboptimal performance. In the fully supervised case, 
\cite{PfisterECCV14} achieves comparable performance as our method. This 
suggests that the method from \cite{BuehlerIJCV11,BuehlerCVPR09} (for hand 
posture modeling) is superior to the one used in our work. We will consider 
combining our relations-based method, for gesture-based recognition, with 
the method from \cite{BuehlerIJCV11,BuehlerCVPR09}, for posture-based 
recognition, as future work. We expect this will improve the precision of 
the posture module which affects the combination of responses; especially 
in cases where signs have similar gestures but slightly different postures.
In addition, different from our method, \cite{FernandoAlCVPR15} considers 
neither relations between parts of the body nor hand posture information. 
In \cite{FernandoAlCVPR15}, hand joints are normalized w.r.t. the torso 
location. This shows that using ranking machines is indeed a powerful 
mechanism for modeling the dynamics of the gestures.
These observations show a strong potential on the combination of advanced 
methods for hand posture modeling \cite{BuehlerCVPR09}, powerful mechanisms 
to model the dynamic of hand gestures \cite{FernandoAlCVPR15} and the 
more detailed relational descriptions proposed in this work, for the task 
of sign language recognition.

% MSR Action3D dataset
For the case of the MSR Action3D dataset, our method has $~3$ pp superior
accuracy compared to the method from \cite{WangCVPR12} which uses a linear 
combination of mined actionlets which are conjunction of the features 
from a subset of the joints of the body. Our method based on the linear 
combination of the postures and gestures responses is comparable to the 
method from \cite{HusseinIJCAI2013} where they use a more expensive 
covariance descriptor to relate the body joints. However, our method 
based on probabilistic combination of the responses produces an improvement 
of $\sim2$ over \cite{HusseinIJCAI2013}. Even though our method is 
designed for sign language recognition, it has a comparable performance 
(~3 pp lower) to the method recently proposed in \cite{PonceBMVC2015} on 
the task of general action recognition in this dataset. Compared to the 
method from \cite{PonceBMVC2015}, our method does not require multiple 
evolution of its models. However, given that \cite{PonceBMVC2015} achieves 
a performance of 0.95 accuracy from a baseline of 0.71, it would be 
interesting to investigate the performance that can be achieved by 
our method (baseline accuracy: 0.92) when integrating such evolutionary 
steps to its models.

% MSRC-12 dataset
On MSRC-12, our method achieves 3 pp over the accuracy reported in 
\cite{EllisIJCV2013} which is focused on a feature vector of pairwise 
joint distances between frames. Furthermore, in this dataset we observe 
a similar trend as in the MSR Action3D dataset where our method is 
slightly superior to the method proposed in \cite{HusseinIJCAI2013}.

\subsection{Computation Time}
\label{sec:exp:compTime}

\begin{figure}
	\centering
		\includegraphics[width=.5\textwidth]{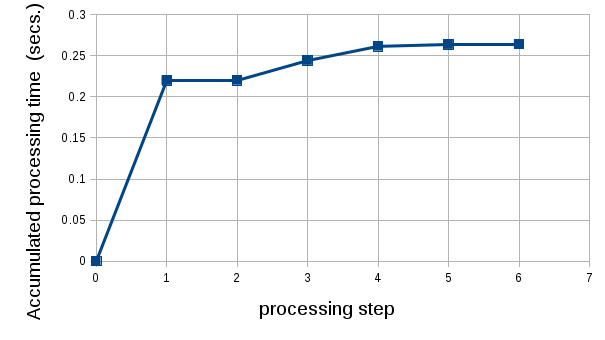}
		\caption{Average Processing times.}
% % % \vspace*{-0.3cm}
\label{fig:processingTimes}
\end{figure}

In order to verify the potential of the proposed method for interactive 
applications, we computed the average processing time during inference.
Our experiments were performed on a single core 2.2 GHZ CPU computer 
with 8 GB of RAM using un-optimized Matlab code.
We summarize the processing times, in seconds, of the different 
stages of our method in Table~\ref{table:processingTimes} and 
Figure~\ref{fig:processingTimes}.

As can be seen in Table~\ref{table:processingTimes}, if done sequentially, 
inference has an average runtime of 0.26415 seconds from which 0.21993 seconds 
are spent on the computation of the posture descriptor. 
Since the focus of this work is on the gesture part, the posture 
module can be improved in future work by faster, and more effective, methods for 
hand posture modeling.

\begin{table}
\caption{Average and accumulated processing times (in seconds) for each 
of the different stages of the proposed method. Notice how the stage related 
to hand postures takes 0.22011 seconds of total time (0.26415 seconds).}
\centering
\begin{tabular}{|l|l|c|c|}
\hline
Stage & Process & Proc. time & Accum. time  \\ \hline

\multirow{2}{*}{Postures} & Descr. Comp.    & 0.21993 & 0.21993  \\ \cline{2-4}
			  & Classification      & 0.00018 & 0.22011  \\ \hline
\hline 			  
	
\multirow{2}{*}{Gestures} & Descr. Comp.    & 0.02417 & 0.24428  \\ \cline{2-4}	 
			  & Classification      & 0.01726 & 0.26154  \\ \hline	 
\hline 			  
	
\multirow{2}{*}{Combination} & Descr. Comp. & 0.00213 & 0.26367  \\ \cline{2-4}	 
			     & Classification   & 0.00048 & 0.26415  \\ \hline	 
\hline 			     
	
\multicolumn{3}{|l|}{Total time} & 0.26415 \\ \hline

\end{tabular}
\label{table:processingTimes}
\end{table}

%%%%%%%%%%%%%%%%%%%%%%%%%%%%%%%%%%%%%%%%%%%%%%%%%%%%%%%%%%%%%%%%%%%%%%%%%%%%%%%%
% % % \section{Acknowledgments}
% % % \label{sec:acknowledgments}
% % % This works is partially supported by FWO project \\G.0.398.11.N.10 ``Multi-camera human behavior monitoring and unusual event detection'',  KU Leuven GOA project CAMETRON, the ``Fondo Europeo de Desarrollo Regional'' (FEDER) and the Spanish
% % % Ministry of Economy and Competitiveness, under its R\&D\&i Support Program in project with ref TEC2013-45492-R.

%%%%%%%%%%%%%%%%%%%%%%%%%%%%%%%%%%%%%%%%%%%%%%%%%%%%%%%%%%%%%%%%%%%%%%%%%%%%%%%%
\section{Conclusion and Future Work}
\label{sec:conclusion}

We presented a method mainly targeted for sign language recognition.
The proposed method focuses on representing each sign by the
combination of responses derived from hand postures and hand gestures.
Our experiments proved that modeling hand gestures by considering 
spatio-temporal relations between different parts of the body brings
improvements over only considering the global trajectories of the hands.
In addition, the proposed method introduces a descriptor for hand 
postures that is flexible to operate on low-resolution images and that
will take advantage of high-resolution images.

Future work will focus on three aspects: 
First, consider state-of-the-art methods to model action dynamics 
to describe the dynamics of hand postures and hand gestures for each 
sign class.
Second, shift the focus of this work towards sign localization/detection.
Third, consider additional features of sign languages such as grammars 
and facial-related gestures. Taking into consideration 
these other characteristics will permit the proposed system to 
develop into a more realistic sign language recognition system.

% \bibliographystyle{spmpsci}      % mathematics and physical sciences
% \bibliographystyle{spphys}       % APS-like style for physics

\bibliographystyle{plain}
\bibliography{paper}   % name your BibTeX data base

% % % % % Non-BibTeX users please use
% % % % \begin{thebibliography}{}
% % % % %
% % % % % and use \bibitem to create references. Consult the Instructions
% % % % % for authors for reference list style.
% % % % %
% % % % \bibitem{RefJ}
% % % % % Format for Journal Reference
% % % % Author, Article title, Journal, Volume, page numbers (year)
% % % % % Format for books
% % % % \bibitem{RefB}
% % % % Author, Book title, page numbers. Publisher, place (year)
% % % % % etc
% % % % \end{thebibliography}

\end{document}